# An Intelligent Multi-Agent Recommender System for Human Capacity Building

Vukosi N. Marivate, George Ssali, and Tshilidzi Marwala

*Abstract*— **This paper presents a Multi-Agent approach to the problem of recommending training courses to engineering professionals. The recommendation system is built as a proof of concept and limited to the electrical and mechanical engineering disciplines. Through user modelling and data collection from a survey, collaborative filtering recommendation is implemented using intelligent agents. The agents work together in recommending meaningful training courses and updating the course information. The system uses a users profile and keywords from courses to rank courses. A ranking accuracy for courses of 90% is achieved while flexibility is achieved using an agent that retrieves information autonomously using data mining techniques from websites. This manner of recommendation is scalable and adaptable. Further improvements can be made using clustering and recording user feedback.**

## I. INTRODUCTION

RECOMMENDER systems, for advising or recommending subject matter, are emerging as a growing application and research field [1]. These systems help users deal with information overload [1]. Information overload results from the availability of a large number of information on any topic in this age. Recommender systems form decision support systems for individuals as well as whole organisations by customising and recommending information. Recommender systems use computational intelligence and can be trained to find patterns between different users and subject matter. In this paper the problem of recommending training courses for engineering professionals is tackled.

This paper presents a system that recommends training courses users by reducing the information overload by using information customisation as well as recommendation. The system further reduces the need for users to search a wide variety of sources of information as well as any need to monitor these sources for changes and updates. The scope of recommendation has been reduced to Electrical and Mechanical engineering professionals. Firstly the background is sketched in section II. Section III reviews existing solutions, section IV is the system overview and section V and VI discuss the agents involved in greater detail. The user interface and details on integration is discussed in section VII. An evaluation of the system is given in section VIII and finally a conclusion is given in section IX.

## II. BACKGROUND

### A. The Recommendation Problem

The problem that most engineering companies face is trying to keep their employees skills updated and thus building their human capacity. To solve this employees are sent to training courses throughout their careers. Thus some companies have complete departments that deal with training. These departments take employee information and use it to look for courses from service providers. The departments often face the problem of not having enough training courses while the employees often feel that the courses offered will not benefit them. When the department or employees search for courses they are then met with numerous results thus are faced with information overload. An intelligent system that can use data from users, service providers and expertise from the training departments and advisors can improve the experiences that the employees have.

### B. Intelligent Agents and Multiagent Systems

Agents are defined by Wooldridge [2] as computer systems that are situated in some environment and capable of *autonomous action* in this environment in order to meet its design objectives. Intelligent agents [2] [3] are defined as agents that can *react* to changes in their environment, have *social ability* (communication) and the ability to use computational intelligence to reach their goals by being *proactive*. Agents are active, task-oriented – modeled to perform specific tasks – and are capable of autonomous action and decision making. Objects on the other hand are passive and noun-oriented – modeled to represent things. So the agent modeling paradigm can be looked at as a stronger encapsulation of localized computational units that perform specific tasks whereas objects model real-world "things" with specific attributes.

By combining multiple agents in one system to solve a problem, the resultant system is a Multiagent system (MAS) [2]. These systems are comprised of agents that solve problems that are simple than the overall system. They can communicate with each other and assist each other in achieving larger and more complex goals. Thus problems that software developers had previously thought as being too complex [4] can now be solved, by localising the problem

Manuscript received December 1, 2007. (Write the date on which you submitted your paper for review.)
V. N. Marivate is with the Computational Intelligence Research Group of the School of Electrical and Information Engineering, Private Bag 3, University of the Witwatersrand, Johannesburg, Wits, 2050. (phone: +27723292126; fax: +27866043509; e-mail: vima@vima.co.za).
G. Ssali is with the Computational Intelligence Research Group of the School of Electrical and Information Engineering, Private Bag 3, University of the Witwatersrand, Johannesburg, Wits, 2050. (e-mail: George.Ssali@macquarie.com).
T. Marwala is with the Computational Intelligence Research Group of the School of Electrical and Information Engineering, Private Bag 3, University of the Witwatersrand, Johannesburg, Wits, 2050. (e-mail: t.marwala@ee.wits.ac.za).

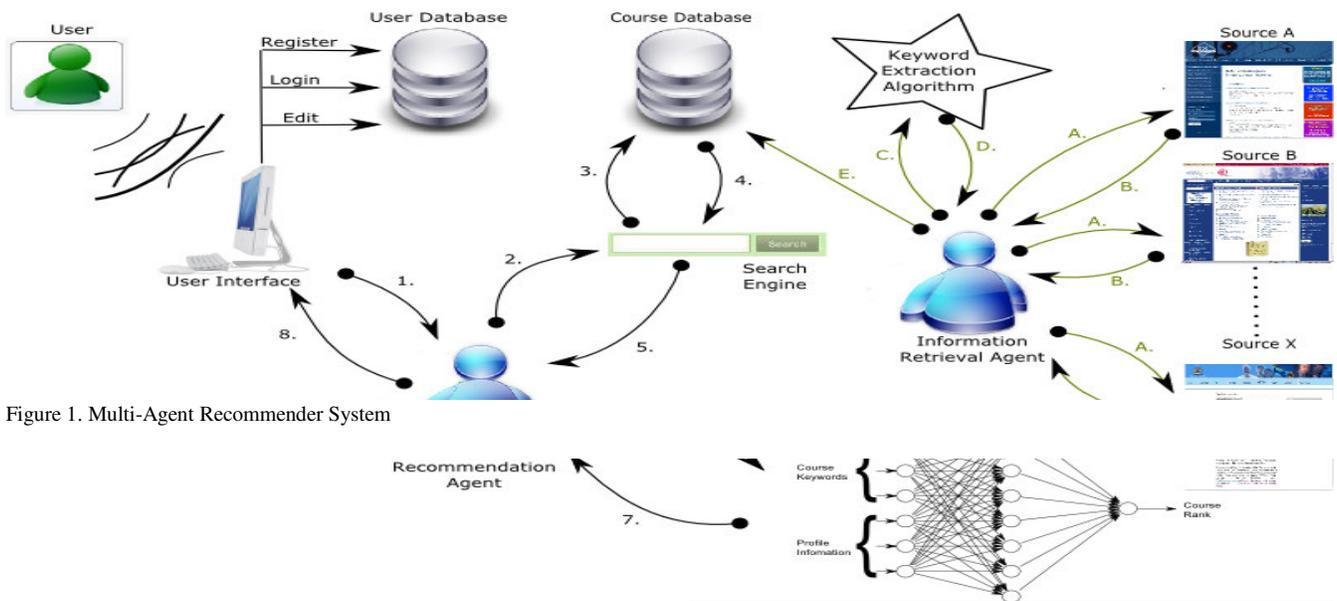

Figure 1. Multi-Agent Recommender System

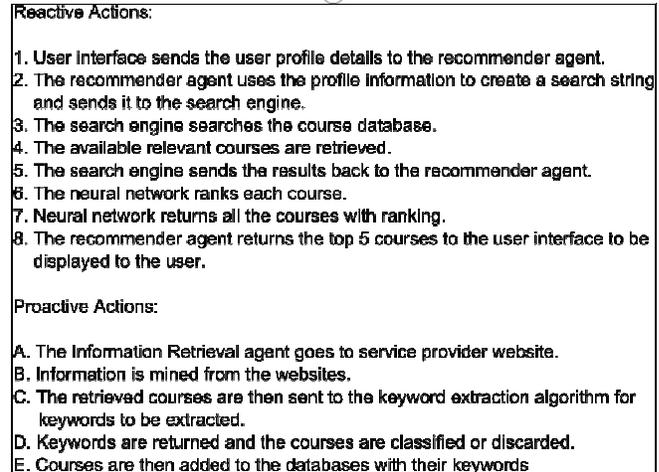

Figure 2. Agent Action Procedures

solving [5]. Multiagent systems have been used in predicting the stock market [6], industrial automation [7] and e-learning systems [8].

### III. REVIEW OF EXISTING SOLUTIONS

There are numerous examples of information customisation systems that have been implemented. Simple examples are search engines [9]; personalised search engines [10] and online stores [11] that use user information to customise adverts and goods being sold. Personalised search engines differ from regular search engines in that they use user habits and preferences [11] to return better search results and recommend other searches that might be useful. Other intelligent recommender systems have also been implemented. An example is MASACAD, which is a Multi Agent System for Academic Advising [12]. It was developed to assist students in choosing courses to take in university. This system had the limitation that given new courses the system had to be retrained. It also is not able to autonomously find new courses as it was only capable of using one source of courses, which is the university database. Autonomous updating is a future problem focus in recommender systems [1] and is solved in the system implemented.

### IV. SYSTEM OVERVIEW

The Multi-Agent solution designed and built for the recommendation problem has two main agents. The first agent is the recommendation agent and the second is the information retrieval agent. The configuration and interactions are shown Figure 1. The User Interface is the gateway to the system for the user, the Recommending Agent proposes a personalized list of training modules, and the Information Retrieval Agent searches a predefined list of service providers' websites for course information and updates. The actions of the agents are described Figure 2 in reference to Figure 1. The proceeding sections describe the agents in further detail.

### V. RECOMMENDATION AGENT

The recommendation agent is a reactive agent that is responsible for using course information as well as user profile information to recommend courses to users using a ranking method. This is done by first searching the course database using the user profile and then ranking each course returned from the search. The recommending agent was built using the Sphinx [13] search engine and the IBM Agent Building and Learning Environment (ABLE) [14].

#### A. User Modelling

To collect useful user information, user modelling had to be carried out [15]. For this to be done properly information in the domain of career guidance and counselling needed to be collected. The users of the system have to be modelled so as to use them in determining what they would consider as good courses. Through consultation with the Career Counselling and Development Unit (CCDU) at the University of the Witwatersrand the user attributes that would be most useful were found. When assisting students with their careers, counsellors at the CCDU look at a number of attributes. The ones chosen for the recommendation

system are:
- Interests (Professional and Personal)
- Abilities (Experience in the chosen field and discipline)
- Goals (Short term and long term)

These attributes then make up the user profile that will be created for each user. The professional interests and discipline are used in the initial search of the course database. The returned results are then sent to the ranking neural network to then rank each course. The experience level and goals of the user were finalised in conjunction with companies Hatch SA and ThyssenKrupp engineering. The experience level was divided into junior, intermediate, senior and management. The goals included professional development, specialisation, marketing etc. The personal interests for example are entertainment, sport, outdoors etc. The personal interests, goals and experience level were used as inputs into the ranking structure. Professional interests depend on the discipline but these can be open pit mining, pumps, energy management etc. This information is stored in a database when a user creates a profile in the system.

### B. Course Information

In order to use courses for recommending, certain information about the course is needed to distinguish courses from each other. Other recommendation systems faced with type of task have resorted to using a fixed number of courses and then training a computational intelligence tool such as a Neural network on these courses only [12]. This way of recommending is not scalable as when new courses are added then the training has to be carried out again. For this system keywords from the courses were used instead to distinguish between courses. Thus when a new course is added to the course database, its keywords are extracted and then used as the course information in the recommendation. This allows for a system that can grow its course list without much retraining or reconfiguration.

From each course a maximum of 3 keywords were extracted. The recommendation system has a keyword extractor that is used by the Information Retrieval agent to extract keywords from the courses that it finds and adds to the course database. Controlled keyword extraction [16] is used so as to make sure that the keywords extracted are those available in our keyword database. The keyword database constructed for the MAS has 233 keywords belonging to the electrical, mechanical or both engineering fields. Each keyword has an id. Thus the keywords resulting from the courses can be used for training the recommender by using their id numbers.

### C. Searching the Database

The course database was populated by finding courses from the different service providers. These courses then needed to be efficiently searched. For this a search engine tool was needed. The Sphinx search engine was used to sift through the databases given search strings. The search strings used in the searches were constructed from the professional interests of the user as well as their discipline. This customised the information returned to the user to their likes. The search engine indexes the course database each time a new course is added and is built in C++ for speed. Speed is rated at an average of 0.1 sec on 2-4 GB of text data. Thus searching does not take a lot of time and results can be processed quickly. After this search the courses are ranked.

### D. Ranking of Courses

To recommend the courses to users ranking was used. This ranking used the course keywords as well as selected profile information. A classification multilayer perceptron neural network was used for the ranking. This neural network configuration is shown in Figure 3.

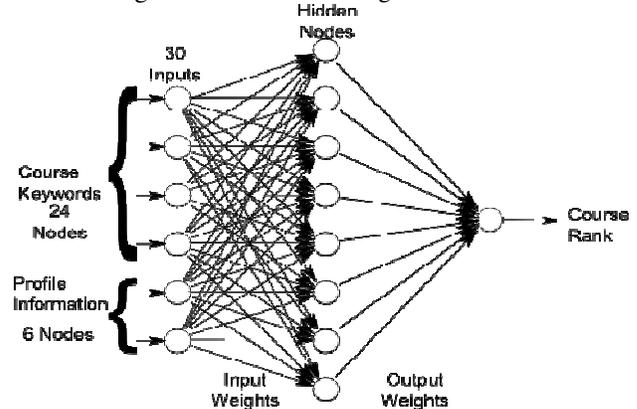

Figure 3. Ranking Neural Network

The general equation of a MLP neural network is shown below (1):

$$y_k = f_{outer}(\sum_{j=1}^{M} w_{kj}^{(2)} * f_{inner}(\sum_{i=1}^{d} w_{ji}^{(1)} x_j + w_{j0}^{(1)}) + w_{k0}^{(2)}) \quad (1)$$

$y_k$ is the rank of the course, $x_j$ represents the inputs into the neural network and $w$ represents the different weights between the nodes in the neural network. For the classification network the $f_{outer}$ activation function is a logarithmic activation function and $f_{inner}$ is the hyperbolic tangent function. Training is done using the back propagation algorithm [17].

*1) Data Collection*

To acquire data to train and test the ranking neural network, a survey was conducted online. Electrical and mechanical engineers from Hatch SA and ThyssenKrupp engineering were requested to fill in the survey. The survey asked the engineers to choose their discipline, professional interests, personal interests and long and short term goals. The engineers were then presented with a list of courses available and then required to choose their best 5. After choosing 5 courses, they then ranked them from 1 (highest) to 5 (lowest). This information was then recorded on our survey database. 250 training data sets were collected from the survey as well as 58 testing data sets. The survey was done in such a manner so as not be biased towards any bias that the companies might have. It was decided that the system should represent the views of the employees, who are the intended users, and not the companies. When asked to put in a section that could be controlled by the employer it was decided it would be best to represent the views of the users. Thus recommendations result from what a similar type of users are interested in. This type of recommendation is collaborative based filtering using regression (NN).

*2) Neural Network Training and Testing*

The data collected from the online survey was used for training the ranking neural network. The inputs into the neural network were the course keywords which were encoded into 8 bit binary numbers thus account for 24 input nodes. The profile information accounted for the rest of the inputs (6) thus resulting in 30 input nodes. Data from the profile is converted to numbers using the key Ids from the database.

Data from the survey is directly accessed using a database import tool from the ABLE toolbox and then sent in as inputs into the neural network. This allowed for rapid use of the data without having to change its format. A typical learning algorithm, back-propagation, compares the NN's output to the desired output, calculates an error, propagates it back through the network and (using a non-linear optimizing algorithm known as "gradient descent") adjusts the NN's weights to decrease this error [18].The numbers of hidden nodes were optimised first. This entails changing the number of hidden nodes in the MLP network and then training. The testing root mean square [19] error is then checked to see how well the network is performing on data it has not seen before. 400 training epochs were used with a learning rate of 0.2 in the ABLE back propagation toolbox. Testing was done with data that was not used to train. The accuracy of the network was measured by the number of correct predictions (rank) within a tolerance of 1 ranking point. Different configurations were tested and are shown in Table 1.

TABLE 1
NEURAL NETWORK CONFIGURATIONS

| Configuration | 32 | 40 | 32 -16 |
|---|---|---|---|
| RMS Error | 0.137 | 0.15 | 0.151 |
| Accuracy (Tolerance of 1) | 90% | 71% | 72% |

32-16 represents a two hidden layer architecture. The network with 32 hidden nodes was chosen as the network to use for ranking. The top 5 – 15 ranked courses would then be displayed to the user. This then would complete the entire recommending process.

## VI. INFORMATION RETRIEVAL AGENT

The information retrieval agent performs two important functions. Firstly it uses data mining to extract courses from service provider websites and then add or update them on the course database. Secondly it extracts keywords from the course description and uses the keywords to classify the course. This agent is based on a simple premise: give it an example of what you want it to retrieve, set it free and wait for it to return the results, and update the course database with any new information from these results. Its task is therefore two-fold: automated data extraction and integration, that is, what to do with the data once it is extracted.

*A. Data Mining*

The vast amount of poorly structured information that exists on public websites makes data extraction a non-trivial task. Furthermore, the content is often scattered over several places on the website, thereby requiring automated navigation like clicking links, filling in forms and crawling to other pages on the site. An inductive machine learning approach was taken to solve this problem. The agent is presented with sample training pages to learn extraction rules which consist of prefix and suffix patterns to mark the beginning and end of target content respectively. Once the rules are learnt, the agent then applies them to extract target items from the other pages [20].

The data extraction was implemented using the scRUBYt framework [21]. This framework, written in Ruby, allows for the use of a fast HTML parser library known as Hpricot [22] as well as the WWW::Mechanize library [23] that effectively provides web browsing functionality in code.

*B. Keyword Extraction and Classification*

The next task of the Information Retrieval Agent: to integrate the extracted data into the existing database, consists of automated keyword extraction and classification of the course data. The keywords are used both to describe the contents of the document and, as previously mentioned, by the Recommender Agent when ranking a given course. Classification on the other hand, improves the relevance of the recommended information that is, electrical engineering courses are not recommended to mechanical engineers and vice versa.

A thesaurus-based keyword extraction algorithm known as KEA++, implemented in Java, was used to perform automated keyword extraction [16]. This machine-learning based algorithm works in two main stages: candidate identification and filtering. Candidate identification involves identifying candidate terms by segmenting the document into tokens on the basis of whitespace and punctuation, followed by the use of a pseudo matching technique [24] to match all n-grams against the controlled vocabulary (thesaurus). A set of attributes is then calculated for each term, the most important being the **TF×IDF** score given by the equation [25]:

$$TD \times IDF = \frac{freq(P,D)}{size(D)} \times -\log_2 \frac{df(P)}{N} \quad (2)$$

where *freq(P,D)* is the number of times *P* occurs in *D*, *size(D)* is the number of words in *D*, *df(P)* is the number of documents containing *P* in the global corpus and *N* the size of the global corpus.

In the filtering stage, a Naive Bayesian classifier model, previously trained on manually indexed course documents, is then used to determine the probability that each word is an index term or not using the formula [25]:

$$P[yes] = \frac{Y}{Y+N} P_{TF \times IDF}[t \mid yes] P_{distance}[f \mid yes] \quad (3)$$

similarly for *P[no]*, where *Y* is the number of positive instances in the training documents, *N* is the number of negative instances (candidate phrases that are not keyphrases), *t* is a feature value derived from Equation 2 above and *f* is the position of the first occurrence of the term. The overall probability that a candidate phrase is a keyphrase is then calculated as:

$$p = \frac{P[yes]}{P[yes] + P[no]} \quad (4)$$

The top ranked candidates are then selected as the document keywords (a more detailed explanation of the algorithm is available in [25]). Once keywords have been extracted, classification is determined via a database look up that maps keywords to the engineering disciplines. Upon completion, the agent then checks to see if the particular course exists and updates the database if it does not.

VII. INTEGRATION AND USER INTERFACE

*A. Integration*

To integrate the agents and the modules such as the keyword extraction and user interface the Symfony [26] web framework was used. The framework allowed for the use of the Model View Control (MVC) paradigm [26] in building a web software application along with Rapid Application Development (RAD) [26]. The MVC paradigm allowed for an object oriented and modular approach for putting the final recommendation system together with a well built user interface. The model is the logic of the application, the view is the interface to the application and the controller connects the two together. This allows for developers to focus on logic without worrying about the look and vice-versa. Thus the resultant complete system was integrated with features such as data security and validation added in without much effort. The parallel building of the interface was due to the MVC paradigm. Programmers need not know about the others modules or actions to build their part of the application. Object relational database [27] use was encouraged by the framework and used. All of the databases used in the recommender system are relational databases [27] built with MySQL database engine. This allows for speed and data integrity through using keys.

*B. User Interface*

The user interface was designed so as to be easy for a user to use. Ease of use in such systems is paramount so that the user need only focus on the task at hand and not on learning how to use the system. Through experience with simple user interfaces, such as that of Google [28], an intuitive interface was built. When the user has registered and logged on to the system he/she is met with their profile information, recommended courses and the search and navigation bar. This is illustrated in Figure 4.

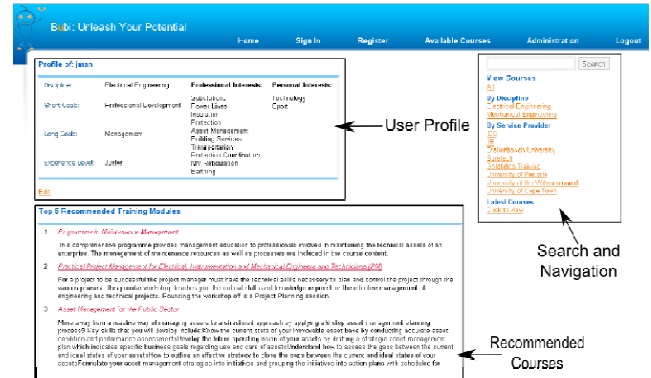

Figure 4. User Interface

There are no complicated actions. The user can simply edit their profile which will automatically then refresh the recommended courses. The user can also search the course database themselves and view course information by service provider or discipline. The Symfony framework encourages the use of logic based programming as well making available tools to validate user actions. If there are errors in data entry then the system warns the user. Users are also not able to access other user's information. The administration module for the user interface was setup to manage users, courses as well as the information retrieval agent's configurations.

VIII. EVALUATION AND DISCUSSION

*A. Evaluation*

The initial goals for the project were to build functioning agents, have a functioning system that could be used for recommendation, a system that is easy to use by the target user that is stable and robust, and a system that is scalable and adaptable. The built system achieved these goals. The recommendation agent has a ranking accuracy of 90% with a tolerance of 1 ranking point as well as a fast search engine integrated. The information retrieval agent is able to go to websites and extract course information as well as classify these courses using a keyword extraction algorithm. The user interface was built to be easy to use and without any complicated procedures. User feedback during open day indicated that this goal was achieved.

The final system is highly decoupled. This is shown by the use of different programming languages used in the implementation. The best language was used for the task at hand e.g. Ruby for Information Retrieval, C++ for the search engine. Thus if a different ranking algorithm needed to be used it can easily be replaced as the framework allows for it. The rules of communication within the agent are the only attributes that need to be kept. Thus the system is scalable as well as being adaptable. E.g. for adaptability, only a change in the user modelling and the courses/subject matter being investigated is needed. Thus the built system can be adapted to problem fields such as job searches, academic advising, business support systems etc. The system cost is low as all of the tools are open source or free to use.

*B. Analysis and Recommendations for Further Work*

The limitation of the built system is the use of the supervised learning neural network for ranking. Thus if a new discipline need be added then the neural network will need to be retrained with new survey data. To remedy this problem online learning methods [29] can be used as well as modifying the current system to be able to provide user feedback. Thus as more users use the system the system can perform better. This would be termed content based filtering recommendation. User feedback in terms of ratings of courses attended can be used as an added input into the system [29].

Due to time constraints keyword and document clustering could not implemented. This would offer better classification of courses [30] as well as offer insight into how multidisciplinary courses are related.

IX. CONCLUSION

The Multiagent system approach for solving the training course recommendation problem is successful in reducing the information overload while recommending relevant courses to users. The system achieves high accuracy in ranking using user information and course information. The final system is scalable and has possibilities for future modification and adaptability to other problem domains. Improvements to the system can be made and the system forms a good platform for future research into the use of computational intelligence in recommender systems.

ACKNOWLEDGMENT

The author would like to thank Raj Naran from Wits CCDU for his input on user modelling. The author would Leon Viljoen from Hatch South Africa, Peter Harris from ThyssenKrupp Engineering and all of the engineers that took part in the online Survey for their assistance.